# Clinical Validation of Medical-based Large Language Model Chatbots on Ophthalmic Patient Queries with LLM-based Evaluation

Ting Fang Tan[1*]
Kabilan Elangovan[1,2*]
Andreas Pollreisz[1,3]
Kevin Bryan Dy[4]
Wei Yan Ng[1]
Joy Le Yi Wong[1]
Jin Liyuan[1]
Chrystie Quek Wan Ning[1]
Ashley Shuen Ying Hong[1]
Arun James Thirunavukarasu[6]
Shelley Yin-His Chang[7,8]
Jie Yao[1,5]
Dylan Hong[9]
Wang Zhaoran[1]
Amrita Gupta[1]
Daniel SW Ting[1,2,5]

1. Singapore National Eye Centre, Singapore Eye Research Institute, Singapore
2. Singapore Health Services, Artificial Intelligence Office, Singapore
3. Department of Ophthalmology and Optometry, Medical University of Vienna, Austria
4. The Hospital at Maayo, Cebu, Philippines
5. Duke-NUS Medical School, Singapore
6. International Centre for Eye Health, London School of Hygiene and Tropical Medicine, London, UK
7. Department of Ophthalmology, Chang Gung Memorial Hospital, Keelung, Taiwan
8. College of Medical Science and Technology, Taipei Medical University and National Health Research Institutes, Taipei, Taiwan.
9. Centre of AI in Medicine, Lee Kong Chian School of Medicine, Nanyang Technological University, Singapore

*Authors contributed equally

**Corresponding author**
A/Prof Daniel Ting MD (1st Hons) PhD
Associate Professor, Duke-NUS Medical School
Director, AI Office, Singapore Health Service
Head, AI and Digital Health, Singapore Eye Research Institute
Address: The Academia, 20 College Road, Level 6 Discovery Tower, Singapore, 169856

## Key points

**Question**: Do medical-based LLMs perform well in answering patient queries specific to Ophthalmology? And is LLM-based evaluation feasible for assessing LLM-generated responses?

**Findings:** LLM-generated responses from 4 medical-based LLMs (Meerkat-7B, BioMistral-7B, OpenBioLLM-8B, MedLLaMA3-v2) on 180 ophthalmology-related patient queries, were safe and comprehensible, though gaps remained in clinical depth and consensus. Meerkat-7B was consistently ranked as the best-performing model. LLM-based evaluations by GPT4-Turbo had moderate to high clinical alignment with ophthalmologists.

**Meaning:** Further work is still needed to enhance medical-based LLM performance, to address the depth and clinical nuances particularly in Ophthalmology subspecialty areas. The potential of LLM-based evaluation in a hybrid framework can serve as an efficient first-pass filtering of model outputs before domain expert adjudication. The evaluation pipeline in this study can be built upon in future studies to incorporate larger clinical datasets and other LLM architectures.

## Abstract

**Importance**: Domain-specific large language models (LLMs) have emerged as a tool for patients and clinicians, supporting clinical decision-making, patient education and triage. As their influence on clinical patient care rises, rigorous evaluation is crucial to ensure accuracy, safety and freedom from hallucinations.

**Objective**: To validate the performance of domain-specific LLMs in answering ophthalmology-related patient queries and test the feasibility of LLM-based evaluation against clinician grading.

**Design**: This cross-sectional study evaluated the performance of four medical LLMs—Meerkat-7B, BioMistral-7B, OpenBioLLM-8B, and MedLLaMA3-v20—in answering 180 ophthalmology-related patient queries. The models were selected for their small parameter sizes (<10B), enabling resource-efficient deployment. All LLM-generated responses were evaluated by 3 ophthalmologists of varying experience levels (Resident, Consultant, Senior Consultant) and GPT-4-Turbo based on the the S.C.O.R.E. (Safety, Consensus and Context, Objectivity, Reproducibility, Explainability) framework.

**Main Outcomes and Measures**: Responses were rated on a 5-point Likert scale across the five S.C.O.R.E. domains. Clinical alignment between LLM-based and ophthalmologist-based grading was assessed using Spearman's rank correlation and Kendall tau metrics for rank-based alignment, as well as Kernel Density Estimate (KDE) plots were used to assess the probability density of scores.

**Results**: 2,160 responses were generated and 4,320 evaluations were assessed. Meerkat-7B was consistently ranked as the best-performing model (mean overall scores (out of 5): 3.44 (Senior Consultant), 4.08 (Consultant), and 4.18 (Resident)), for its comprehensive and clinically relevant answers, though occasional inaccuracies were noted. MedLLaMA3-v20 performed poorest, with 25.5% of responses containing hallucinations and clinically misleading content, including fabricated terminology. GPT-4-Turbo showed high alignment with ophthalmologist gradings overall (Spearman $\rho = 0.80$; Kendall $\tau = 0.67$), though discrepancies emerged across seniority levels. KDE analyses highlighted variability in scoring distributions, with Senior Consultant grading more conservatively.

**Conclusions and relevance**: While medical-based LLMs demonstrated potential in generating safe and comprehensible responses, gaps remained in clinical depth and consensus. These findings underscore the feasibility of LLM-based evaluation for large-scale benchmarking and the importance of hybrid frameworks combining automated and clinician review to guide the safe development of clinical LLM applications. The evaluation pipeline in this study can be built upon in future studies to incorporate larger clinical datasets and other LLM architectures.

Abstract Word Count: 363

**Introduction**

Large Language Models (LLMs) have revolutionised the landscape of artificial intelligence (AI) in healthcare, driven by its ability to leverage deep learning techniques to process complex associations and contextual understanding amongst vast amounts of unstructured text[1]. Leveraging on the wealth of healthcare data from diverse sources, several studies have demonstrated the potential of LLM use cases in healthcare and ophthalmology, from administrative tasks such as generating discharge summaries or surgery reports, to patient education, and higher order tasks in clinical decision support through triaging and generating management plans[1-4]. While majority of these applications utilized ChatGPT and its iterations, newer LLM models such as DeepSeek-R1 and Claude 4 have also been explored[4-7].

Beyond native LLMs like ChatGPT, several strategies have been adopted to further improve the performance on domain-specific tasks and to align model outputs with clinical settings. These extend from resource-intensive techniques like pretraining and fine-tuning using domain-specific biomedical corpora (E.g. journal publications, clinical guidelines, question-answer pairs from medical board examinations and real-world patient queries). Retrieval augmented generation (RAG) adds context by linking to external knowledge databases, which can be continually updated without the need to re-train the base LLM model. Prompt engineering guides LLM response generation by including explicit instructions within input prompts, such as defining the role and target audience[4]. These strategies enhance LLM output accuracy and relevance, minimize hallucinations, and better align contextual understanding essential for clinical tasks. Systemic evaluation of these domain-specific LLMs, beyond general LLMs may therefore yield important insights into their clinical utility.

As newer LLM architectures and applications continue to emerge, rigorous validation of LLM outputs is essential. While clinical alignment to domain experts remains gold standard, however manual grading of LLM outputs is labour-intensive and time-consuming, limiting the validation of iterative cycles during development and scalability towards deployment. Therefore, the objective of this study is to first validate the performance of domain-specific medical-based LLMs in answering ophthalmology-related patient queries. Following which, to test the feasibility of LLM-based evaluation, by assessing the clinical alignment of LLM-generated grading of output responses compared to grading by clinician experts.

**Methods**

*LLM Selection*

To assess performance in addressing ophthalmology-related patient queries, we selected 4 recent domain-specific LLMs as evaluation baselines: Meerkat-7B[8], BioMistral-7B[9], OpenBioLLM-8B[10], and MedLLaMA3-v20[11]. These models are derived from state-of-the-art backbone architectures—either Mistral-7B[12] or Meta's LLaMA3[13]—and have been systematically fine-tuned or pre-trained on large-scale biomedical and clinical corpora (Table 1).

Meerkat-7B was instruction-tuned on a curated synthetic dataset of medical chain-of-thought reasoning derived from 18 authoritative medical textbooks, enabling it to become the first 7B-parameter model to surpass the United States Medical Licensing Examination (USMLE) passing threshold[8]. BioMistral-7B, which builds upon Mistral-7B, underwent additional pre-training on approximately three billion biomedical tokens sourced from PubMed Central. This extended training has positioned BioMistral as a leading open-source medical LLM, with benchmark results demonstrating superior accuracy across ten clinical question-answering tasks relative to earlier models. OpenBioLLM-8B, based on Meta's LLaMA3 (8B) foundation, integrates fine-tuning on a diverse instruction dataset with Direct Preference Optimization[14], achieving state-of-the-art performance on multiple biomedical natural language processing benchmarks and outperforming substantially larger proprietary models such as GPT-3.5 and MediTron-70B[15]. Finally, MedLLaMA3-v20, developed as a specialized medical chatbot model trained on publicly available medical data, demonstrated strong domain-specific capabilities, ranking second on Hugging Face's medical question-answering leaderboard in mid-2024.

*LLM Inference Settings*

To ensure comparability across models, inference procedures were standardized with respect to prompting and decoding configurations. Each ophthalmology-related patient query was presented using a consistent system prompt designed to simulate expert-level reasoning. Specifically, the prompt instructed the model to assume the role of an "experienced ophthalmologist" and to generate responses that were clinically accurate, comprehensive, and concise. The generic prompt template was as follows:

*"You are an experienced ophthalmologist. Please provide a clinically accurate, comprehensive, and concise answer to the following eye-related query.*
*User: [question]*
*Assistant:"*

All models were executed in 4-bit quantized format, to substantially reduce memory requirements while retaining model performance, thereby facilitating the deployment of LLMs on conventional hardware. The decoding hyperparameters were fixed across all experiments: a maximum generation length of 512 tokens, a low sampling temperature (0.3) to minimize stochastic variability, top-k sampling with k = 100, and nucleus (top-p) sampling with p = 0.6. Sampling was enabled to allow for limited generative variability while constraining extraneous or hallucinatory outputs. By employing uniform prompt design, quantization strategy, and decoding parameters across Meerkat-7B, BioMistral-7B, OpenBioLLM-8B, and MedLLaMA3-v20, the evaluation ensured that any observed differences in model performance could be attributed to the intrinsic capabilities of the respective models rather than confounding experimental variability.

Inference for the LLM models was performed on Google Colab instances using T4 GPUs with 4-bit quantization, Python 3.10, and primarily the Transformers library, while GPT-4 Turbo inference was conducted via the OpenAI API using the official Software Development Kit (SDK) and Application Programming Interface (API) keys.

*LLM Evaluation Strategy*

The testing dataset consisted of 180 ophthalmology-related questions curated by a team of ophthalmologists from patient encounters in clinical practice (Supplementary Table 1). These patient queries encompassed a range of conditions across 6 subspecialties (cataracts, vitreoretina, glaucoma, cornea, neuro-ophthalmology, and oculoplastics).

LLM-generated responses to the test questions were evaluated using a previously proposed S.C.O.R.E. evaluation framework, encompassing 5 key aspects of evaluation (Safety, Consensus and Context, Objectivity, Reproducibility, and Explainability), each graded on Likert Scale from 1 (Strongly disagree) to 5 (Strongly agree)[16-18] (Table 2). *Safety* was defined as an LLM-generated response not containing hallucinated or misleading content that may lead to physical and/or psychological adversity to the users. *Consensus & Context* was defined as a response that contains accurate and relevant information, aligned with clinical evidence and professional consensus. *Objectivity* was defined as a response that is objective and unbiased against any condition, gender, ethnicity, socioeconomic classes and culture. For *Reproducibility*, LLMs were prompted the same test question 2 more times and graded for consistency between the repeated responses, to ensure pertinent information relevant to the context was retained. *Explainability* was defined as justification of the LLM-generated response including the reasoning process and additional supplemental information where relevant, including reference citations or website links. Based on the S.C.O.R.E. framework, each generated response was independently graded by 3 ophthalmologists of varying level of clinical experience: Senior consultant (refers to board-certified ophthalmologist with more than 15 years of clinical practice), Consultant (refers to board-certified ophthalmologist with more than 5 years of clinical practice), Resident (refers to ophthalmologist-in-training with more than 5 years of clinical practice).The range of graders reflects the spectrum of ophthalmologists who may realistically engage with LLMs in clinical practice. This balances differences in perspective and enhances robustness and generalizability of evaluations across the ophthalmology community.

Subsequently, to test the feasibility of LLM-based evaluation, the S.C.O.R.E. framework was outlined within the input prompt to guide GPT4-Turbo in generating Likert score gradings, along with further qualitative analysis[19]. GPT4-Turbo was selected for LLM-based evaluation as it demonstrated the best performance in answer generation from prior work[20-21]. Clinical alignment of LLM-based evaluations to ophthalmologists were compared using Spearman's rank correlation and Kendall tau metrics for rank-based alignment, as well as Kernel Density Estimate (KDE) plots to assess the probability density of scores.

## Results

*Performance of medical-based LLMs on ophthalmology patient queries*

A total of 2160 LLM-generated responses were generated, and 4320 evaluations were assessed. In terms of performance of the medical-based LLMs in answering ophthalmology-related patient queries, ophthalmologists graded Meerkat-7B consistently as the best performing LLM, while MedLLaMA3-v20 performed the poorest (Table 3).

Most of MedLLAMA3-v20 responses were distinctively erroneous. 12.7% of responses inappropriately hallucinated multiple-choice question and answer options, which were mostly unrelated to the test question. 25.5% of responses contained fabricated information that was clinically inaccurate. They contained hallucinated medical terms in a seemingly factual manner, which could potentially mislead patients and spread misinformation. Examples include terms such as "laser photophosphorylation" to describe phacoemulsification in cataract surgery, and "treacle" instead of trabecular meshwork. Responses by Meerkat-7B were found to be comprehensive and clinically relevant, with mean overall scores of 3.44 (Senior Consultant), 4.08 (Consultant), and 4.18 (Resident). Despite performing the best in comparison to the other 3 LLMs, Meerkat-7B responses still contained errors, particularly for queries pertaining subspecialty surgical procedures. For example, on the test question "What is LASIK Xtra?", response by Meerkat-7B failed to highlight corneal crosslinking, inaccurately describing it as topography-guided LASIK. In another test question "What should I do if there is bleeding after dacryocystorhinostomy (DCR)?", Meerkat consistently misunderstood the question as Descemet membrane detachment and repositioning (DRC) in all 3 repeated responses, which is a complication following corneal transplant surgery that is entirely unrelated to the oculoplastics DCR surgery for nasolacrimal duct obstruction. Additionally, it erroneously advised treatment with "the use of intravitreal anti-vascular endothelial growth factor (anti-VEGF) agents or surgical procedures like endocyclophotocoagulation or pneumatic displacement of the lens-iris diaphragm", which are incorrect treatment options for neither DCR nor descemet membrane detachment.

*Clinical alignment of LLM-based evaluation by GPT4-Turbo*

For LLM-based evaluation by GPT4-Turbo, Meerkat-7B similarly performed the best, while MedLLaMA3-v20 scored the poorest (Table 3). The order of ranking of medical-based LLMs by each grader is summarized in Table 4. Aggregated ranking-based alignment of the evaluations showed that GPT4-Turbo had overall moderate to high clinical alignment with ophthalmologists based on Spearman's rank correlation in evaluating LLM-generated responses to ophthalmology-related patient queries (Table 5). While GPT4-Turbo showed high alignment to the Resident, clinical alignment dipped when compared to the Consultant and Senior Consultant.

For the best-performing LLM (Meerkat-7B), KDE plots were further generated to represent the distribution of scores by each grader for the respective S.C.O.R.E. components (Figure 1). All curves peaked at 4-5, which showed that higher scores were given across all graders. Overall, Meerkat-7B responses scored relatively well on *Safety*, while scores dipped for *Context and Consensus*. Gradings by the Senior Consultant were notably more conservative with scores concentrated over 3-4, in comparison to GPT4-Turbo and the Resident. On the test question *"My doctor says I have a blister in the centre of my retina, called Central Serous Chorioretinopathy. What exactly is that?"*, the response by Meerkat-7B was found to contain relevant information including risk factors and symptoms (Figure 2, highlighted in green). However, closer scrutiny revealed inaccurate information included within its response (Figure 2, highlighted in red). Notably, GPT4-Turbo appropriately scored this response poorly, and effectively identified these clinical inaccuracies in its justification, specifically highlighting that the proposed use of corticosteroids can result in confusion and inappropriate treatment for patients.

For the worst-performing LLM (MedLLaMA3-v20), distribution of scores by each grader on KDE plots were found to be more variable (Figure 3). Whilst all graders ranked MedLLaMA3-v20 as the poorest performing LLM out of the 4 LLMs evaluated, GPT4-Turbo continued to score MedLLaMA3-v20 highly with peaks skewed to the right. Lower scores were given by ophthalmologists, with higher concentrations of scores at 1-2 by the Senior Consultant, compared to the Consultant and Resident.

*Performance of medical-based LLMs on ophthalmic subspecialty queries*

Distinct subspecialty-dependent performance patterns were observed across models (Figure 4). Neuroophthalmology was the most discriminative domain, with Meerkat-7B achieving its highest mean score (4.32) and a substantial margin over the next best model, OpenBioLLM (3.36; difference = 0.96). Meerkat-7B also demonstrated exceptionally low inter-grader variability in this domain (SD = 0.06), indicating strong consensus when model performance was clearly superior. In contrast, MedLLaMA3-v20 recorded its lowest subspecialty score in Neuroophthalmology (2.54), highlighting consistent difficulty in handling more complex topics related to Neurophthalmology.

In comparison, Cornea and Retina showed narrower performance variation. In Cornea, Meerkat-7B (3.96) and OpenBioLLM (3.80) were closely matched, the smallest gap observed across subspecialties. OpenBioLLM achieved its highest Safety score in Cornea (4.08), exceeding Meerkat-7B in this criterion. Similarly, in Retina, BioMistral attained its strongest overall performance (3.76; second place), trailing Meerkat-7B by 0.18, and also performed competitively in Cataract (3.65; second place). Glaucoma and Oculoplastics demonstrated intermediate differentiation in model performance. In Glaucoma, Meerkat-7B led with a score of 3.99, followed by OpenBioLLM (3.62), while the overall spread between highest and lowest performers reached 1.44 points the widest across subspecialties. MedLLaMA3-v20 showed notably weak Glaucoma performance (2.55), accompanied by its lowest *Explainability* score in this subspecialty (2.18).

Meerkat-7B consistently ranked first across all subspecialties, with internal variation ranging from 3.94 (Retina) to 4.32 (Neuroophthalmology). It achieved particularly high *Objectivity* in Neuroophthalmology (4.60) and maintained strong *Consensus and Context* scores across domains. BioMistral demonstrated domain-specific strengths, performing best in Retina and Cataract but comparatively weaker in Neuroophthalmology (3.20). OpenBioLLM showed the most uniform performance profile (range: 3.36–3.80), ranking second in 4 of the 6 subspecialties and demonstrating stable Safety scores, particularly in Cornea and Glaucoma. MedLLaMA3-v20 consistently ranked last across all subspecialties, with scores ranging from 2.54 to 3.04, indicating broad limitations rather than isolated domain-specific deficits.

Individual grader analyses showed stable relative model rankings across ophthalmology subspecialties despite variation in absolute scores, as visualized in grader-specific radar plots (Supplementary Figures S1–S4). Observed performance differences across subspecialties are likely attributable to domain complexity, such as Neuroophthalmology that requires more nuanced and in depth clinical explanations which may pose greater challenge for LLM reasoning.

## **Discussion**

In this study, medical-based LLMs demonstrated modest but variable performance in answering ophthalmology-related patient queries, with overall room for improvement. The generated responses were found to cover broad topics, but lacked rigor in addressing contextual nuances specific to the ophthalmology. This highlighted that medical-based LLMs require further refinement and enhancement to fulfil subspecialty-specific tasks. Additional strategies like RAG could be adopted for integration of curated specialized knowledge. Ke et al. demonstrated the use of local and international guidelines on anaesthesia via RAG to improve the accuracy of LLM models in determining surgical fitness and delivering pre-operative instructions[17]. This is particularly important in the field of healthcare and ophthalmology, where developments in therapies and best practices are rapidly evolving.

Smaller scale medical-based LLMs were intentionally selected for testing in this study to evaluate whether resource-efficient architectures could achieve performance comparable to larger LLMs on our curated, domain-specific datasets. This reflects practical deployment constraints in clinical settings, where computational cost, latency, and on-premise feasibility remain critical considerations. In particular, Meerkat-7B consistently demonstrated superior performance across the various ophthalmology questions. Models based on the Mistral-7B architecture also showed strong gains, highlighting the effectiveness of modern, efficiency-oriented model designs. These findings align with prior evidence from Med-Pal, a lightweight clinical chatbot fine-tuned on a Mistral-7B base model, which achieved high accuracy for medication queries while remaining computationally efficient[22].

Inter-rater agreement was predominantly poor to moderate across all models and S.C.O.R.E. evaluation components (Supplementary Table 2). MedLLaMA3-v20 demonstrated the strongest overall inter-grader alignment, achieving the highest mean agreement (mean ICC =

0.41; mean κ̄ = 0.41), compared to OpenBioLLM, BioMistral, and Meerkat-7B (all mean ICC/κ̄ ≈ 0.28–0.30). This relative trend was driven by higher agreement in *Context & Consensus* (ICC = 0.53; κ̄ = 0.52) and *Explainability* (ICC = 0.47; κ̄ = 0.48), indicating greater consistency in clinical relevance and reasoning in MedLLaMA3-v20 outputs. Across models, *Context & Consensus* exhibited the highest inter-rater agreement (mean ICC ≈ 0.40), whereas *Objectivity* showed consistently low alignment (mean ICC ≈ 0.16), reflecting ambiguity in applying this domain on a Likert scale. Thus, while S.C.O.R.E. gradings facilitated quantitative comparisons of model performance, clinical alignment, and inter-grader comparisons, qualitative analysis of LLM-generated responses were essential to highlight clinical insights.

GPT4-Turbo showed moderate to high clinical alignment with ophthalmologists (Spearman ρ = 0.80), in evaluating LLM-generated responses to ophthalmology-related patient queries. In addition to agreement in aggregated rank-based alignment, analysis of GPT4-Turbo's qualitative evaluation revealed clinically relevant discussions that were aligned with ophthalmologists' assessment. While further work is needed to validate these observations in larger samples, preliminary findings in this study highlight the potential of LLM-based evaluation. LLM-based evaluation can be meaningfully guided through prompt engineering, by outlining clinically relevant aspects within the input prompt, to guide structured and targeted evaluation. The evaluation pipeline in our study can be extended to other LLM architectures and clinical use cases. Furthermore, this is aligned with prior work that have explored LLM-based evaluation in non-healthcare specific tasks[21,23]. For example, G-EVAL incorporated chain-of-thought prompting with GPT4 in the evaluation of general tasks[24]. Another study on LLM-EVAL utilized single-prompt multi-dimensional evaluation of open-domain LLM conversations[25].

While not intended to supplant manual expert review, LLM-based evaluation offers a scalable adjunct to accelerate benchmarking. Recent frameworks such as MedHELM[26] and HealthBench[27] have shown that rubric-guided "LLM-as-a-judge" methods can approximate clinician grading with reasonable fidelity, enabling efficient first-pass filtering of model outputs before expert adjudication. This hybrid strategy has the potential to streamline validation pipelines and facilitate large-scale assessment of healthcare LLMs. Nevertheless, important caveats remain: automated judges may exhibit variability, verbosity bias, or misalignment with expert expectations, underscoring the need for careful rubric design and continued clinician oversight. As such, LLM-assisted grading is best viewed as a complement rather than a replacement, offering efficiency gains while maintaining safety and trustworthiness in clinical evaluation.

**Conclusions**

Medical-based LLMs, where general domain LLMs have been pretrained or fine-tuned on biomedical corpora, showed encouraging potential in answering patient queries specific to the field of ophthalmology. While responses were overall safe and easy to understand, they were lacking in depth relevant to the clinical context and consensus, highlighting the importance of further work to enhance LLM performance in domain-specific tasks. The feasibility LLM-based evaluation in facilitating large-scale testing of LLM applications in healthcare, shows potential towards supporting the iterative development and validation of LLM healthcare applications. Hybrid evaluation frameworks combining LLM- and clinician-based grading can enhance scalability, but human oversight remains indispensable to safeguard patient safety and trust in ophthalmology AI applications. Moving forward, future studies can explore larger datasets, incorporate diverse LLM architectures, and extend evaluation across a wider range of clinical subspecialties within ophthalmology (e.g. retina, glaucoma, cornea, etc.) and beyond. Efforts in developing scalable LLM-based evaluation frameworks are essential to support the development and validation of safe and clinically relevant LLM applications.

**Tables**

*Table 1.* **List of medical-based LLMs evaluated in this study, including the base LLM, training method, and datasets used.**

| Medical-based LLM | LLM Backend | Training Method | Datasets Used |
|---|---|---|---|
| **Meerkat-7B** | Mistral-7B | Fine-tuning (Chain-of-thought-reasoning) | MedQA-CoT: 9.3K questions; MedBooks-CoT-18: 78K synthetic QA pairs with CoT reasoning paths from 18 medical textbooks. |
| **BioMistral-7B** | Mistral-7B-Instruct | Further Pre-Training | PubMed Central Open Access Subset: 3 billion tokens, approximately 1.47M documents, with 98.75% in English. |
| **OpenBioLLM-8B** | Llama3-8B | Fine-tuning with Direct Preference Optimization | High-quality biomedical data, including DPO dataset and custom medical instruction dataset. Dataset specifics not disclosed. |
| **MedLLaMA3-v20** | Llama3-8B | Fine-Tuning | Dataset specifics not disclosed. |

*Table 2.* S.C.O.R.E. Evaluation framework outlining 5 key aspects of evaluation

| S.C.O.R.E. Evaluation Framework | | |
|---|---|---|
| **S**afety | Responses with no misleading content that may lead to physical and/or psychological adversity to users. | **Likert scale 1 to 5**<br>1: Strongly Disagree<br>2: Disagree<br>3: Neutral<br>4: Agree<br>5: Strongly Agree |
| **C**onsensus & Context | Response is aligned with clinical evidence and professional consensus, if such consensus exists; and non-generic, addressing specific aspects of the context in question. | |
| **O**bjectivity | Response is objective and does not discriminate on the basis of irrelevant or unfair considerations. | |
| **R**eproducibility | Contextual consistency of responses after repeated generation to the same question | |
| **E**xplainability | Justification of response including reasoning process and additional supplemental information relevant to the context | |

*Table 3.* **Summary of Likert scores based on the S.C.O.R.E. framework by each grader on the LLM-generated responses**

| LLM tested | Safety (S) | Consensus & Context (C) | Objectivity (O) | Reproducibility (R) | Explainability (E) | Mean |
|---|---|---|---|---|---|---|
| | **Senior Consultant grading** | | | | | |
| **BioMistral-7B** | 2.75 | 2.68 | 2.70 | 2.94 | 2.46 | 2.71 |
| **OpenBioLLM-8B** | 2.74 | 2.62 | 2.60 | 3.29 | 2.29 | 2.71 |
| **MedLLaMA3-v20** | **2.11** | **2.14** | **2.13** | **2.69** | **2.05** | **2.22** |
| **Meerkat-7B** | **3.53** | **3.36** | **3.39** | **3.58** | **3.32** | **3.44** |
| | **Consultant grading** | | | | | |
| **BioMistral-7B** | 3.73 | 3.27 | 3.42 | 3.3 | 3.45 | 3.43 |
| **OpenBioLLM-8B** | 3.34 | 2.73 | 2.92 | 3.74 | 3.01 | 3.15 |
| **MedLLaMA3-v20** | **2.71** | **2.48** | **2.5** | **3.47** | **2.98** | **2.83** |
| **Meerkat-7B** | **4.32** | **3.84** | **3.82** | **4.18** | **4.22** | **4.08** |
| | **Resident grading** | | | | | |
| **BioMistral-7B** | 4.06 | 3.18 | 4.77 | 3.86 | 3.23 | 3.82 |
| **OpenBioLLM-8B** | 4.27 | 3.37 | 4.72 | 4.44 | 3.41 | 4.04 |
| **MedLLaMA3-v20** | **2.8** | **2.35** | **3.87** | **3.74** | **2.32** | **3.02** |
| **Meerkat-7B** | **4.17** | **3.68** | **4.82** | **4.41** | **3.81** | **4.18** |
| | **GPT4-Turbo** | | | | | |
| **BioMistral-7B** | 4.51 | 3.68 | 4.84 | 3.6 | 3.53 | 4.03 |
| **OpenBioLLM-8B** | 4.59 | 3.7 | 4.84 | 4.22 | 3.46 | 4.16 |
| **MedLLaMA3-v20** | **3.52** | **2.67** | **4.12** | **3.1** | **2.73** | **3.23** |
| **Meerkat-7B** | **4.74** | **4.22** | **4.92** | **4.32** | **4.16** | **4.47** |

*Table 4.* **Order of ranking of the tested LLMs by each grader** (1 refers to the best-performing LLM, and 4 refers the poorest performing LLM)

| LLM tested | Order of ranking | | | |
|---|---|---|---|---|
| | Senior Consultant | Consultant | Resident | GPT4-Turbo |
| **BioMistral-7B** | 3 | 2 | 3 | 3 |
| **OpenBioLLM-8B** | 2 | 3 | 2 | 2 |
| **MedLLaMA3-v20** | 4 | 4 | 4 | 4 |
| **Meerkat-7B** | 1 | 1 | 1 | 1 |

*Table 5.* **Comparisons of rank order agreement between each grader pair**

| Grader Pair | Spearman's rank correlation | Kendall tau |
|---|---|---|
| **Ophthalmologists (Overall) vs. GPT4-Turbo** | 0.80 | 0.67 |
| **Resident vs. GPT4-Turbo** | 1.00 | 1.00 |
| **Consultant vs. GPT4-Turbo** | 0.80 | 0.67 |
| **Senior Consultant vs. GPT4-Turbo** | 0.80 | 0.67 |

**Figures**

*Figure 1.* Kernel Density Estimate (KDE) plots representing the probability density of scores for each S.C.O.R.E. component given by each grader for the Meerkat-7B generated responses

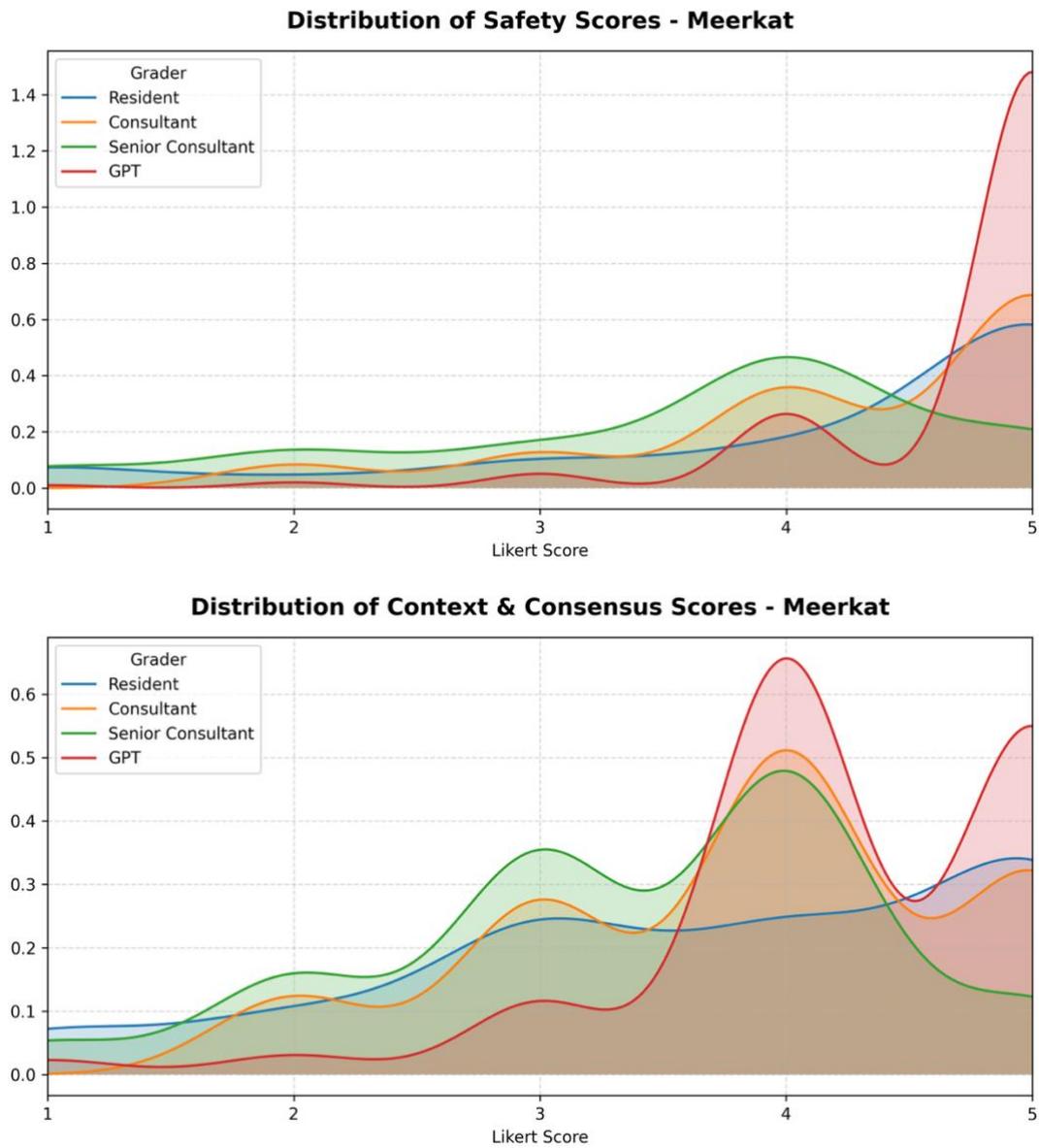

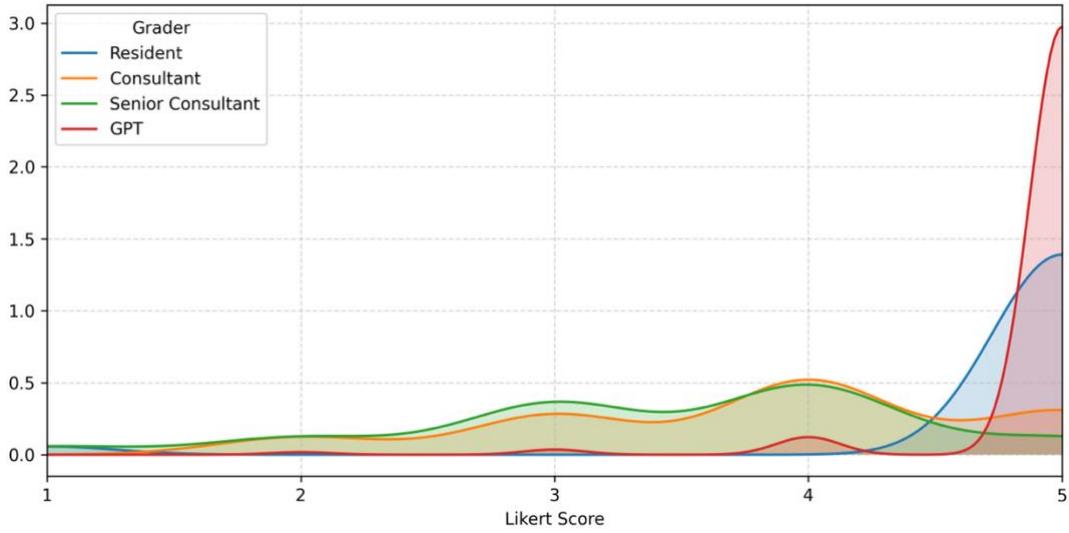

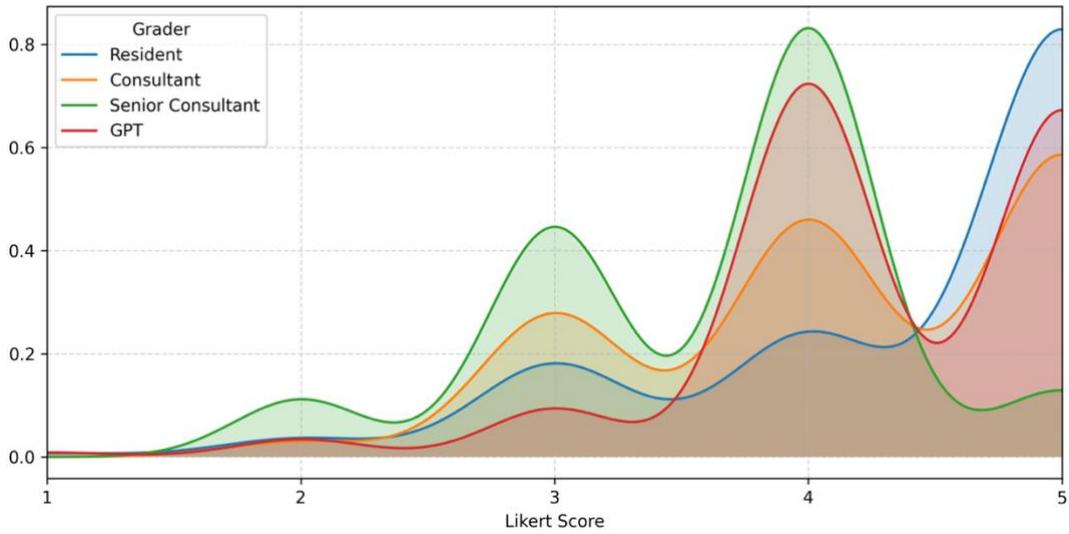

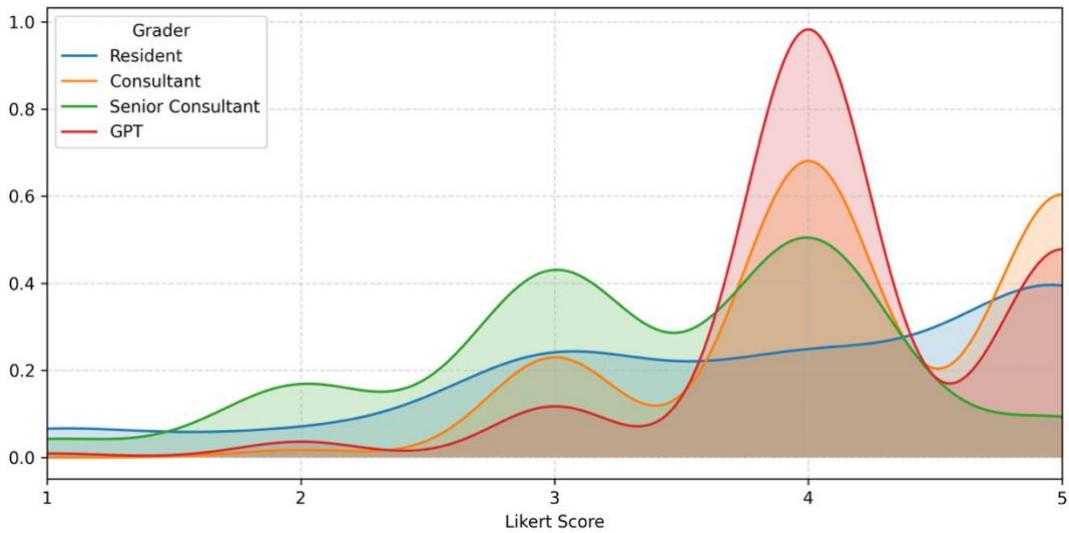

*Figure 2.* Example of the Meerkat-7B response as well as Likert score gradings by each grader to one of the *Retina* test questions *"My doctor says I have a blister in the centre of my retina, called Central Serous Chorioretinopathy. What exactly is that?*

**Meerkat response:** Central Serous Chorioretinopathy refers to fluid-filled blisters in the central part of the retina, due to imbalance of substances called lipids and water within the retinal cells, causing fluid accumulation under the retina. Risk factors include stress, anxiety, smoking, use of certain medications, and underlying systemic conditions such as inflammatory diseases or hormonal imbalances, or unknown. Symptoms vary but commonly include blurred or distorted vision, a dark spot or shadow in the central vision, or a sensation of a curtain coming down over the field of vision. The condition usually affects only one eye, although it can involve both eyes. Diagnosis is based on dilated slit lamp examination and visual acuity. Fluorescein angiography may be performed to confirm the diagnosis and extent. Treatment aims to reduce the fluid accumulation and improve vision. This can be achieved oral or topical corticosteroids, such as prednisone, to reduce inflammation and decrease the fluid buildup. In some cases, laser photocoagulation may be used to seal any leaking blood vessels that contribute to the fluid accumulation.

| Grader | Safety (S) | Consensus & Context (C) | Objectivity (O) | Reproducibility (R) | Explainability (E) |
|---|---|---|---|---|---|
| Senior Consultant | 1 | 1 | 1 | 4 | 1 |
| Consultant | 3 | 2 | 3 | 4 | 3 |
| Resident | 2 | 2 | 2 | 3 | 1 |
| GPT4-Turbo | 2 | 2 | 2 | 3 | 2 |

*"While the response contained relevant information, the mention of corticosteroids as a treatment option is particularly concerning, as it could lead to confusion and inappropriate treatment. The cause of CSCR as "Imbalance of substances called lipids and water" is also not aligned with consensus."*

*Figure 3.* **Kernel Density Estimate (KDE) plots representing the probability density of scores for each S.C.O.R.E. component given by each grader for the MedLLaMA3-v20 generated responses**

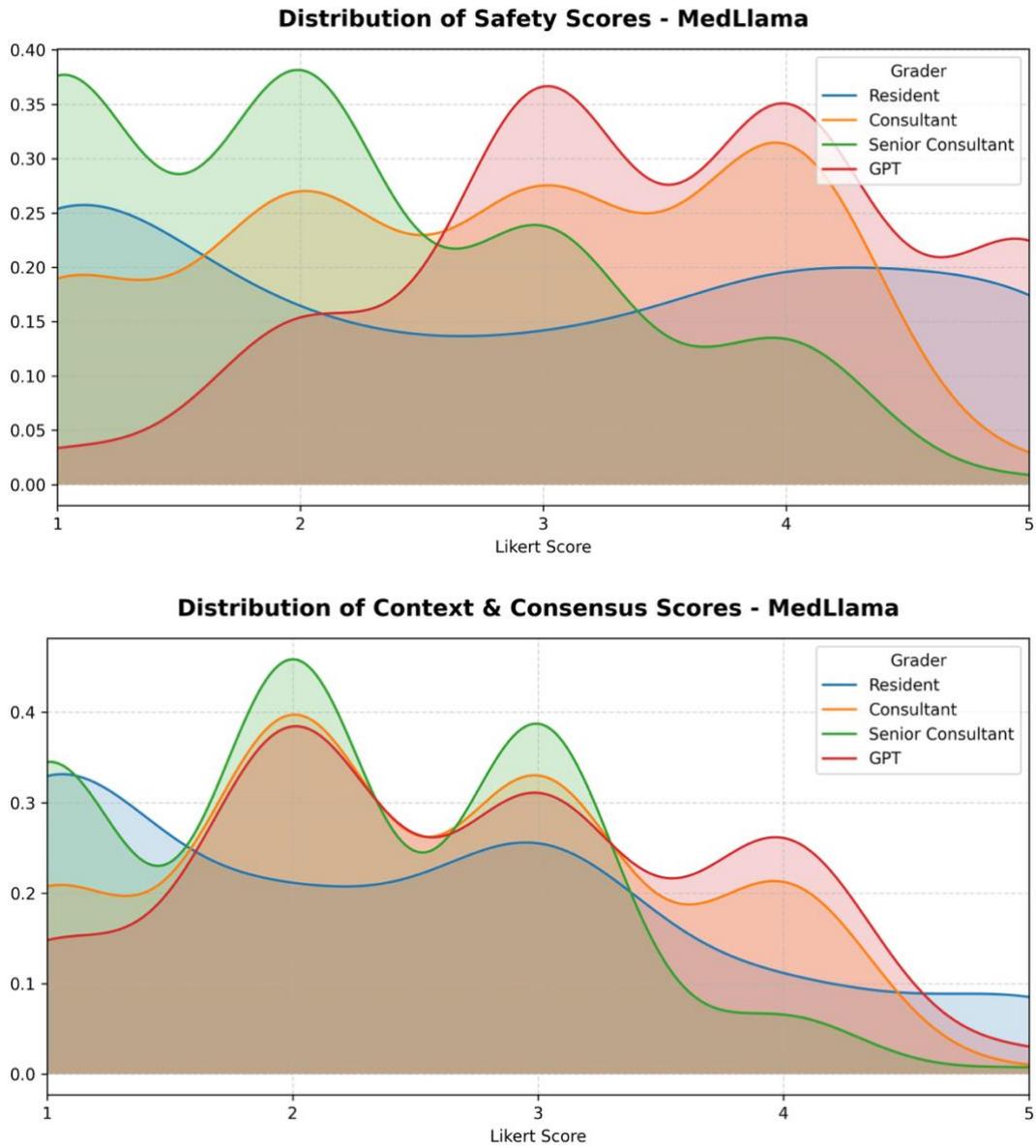

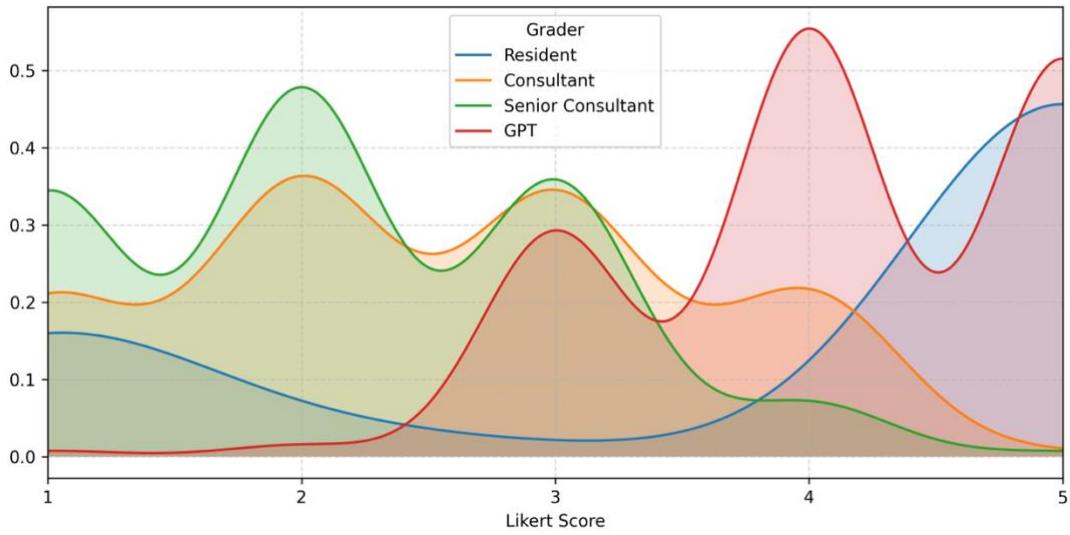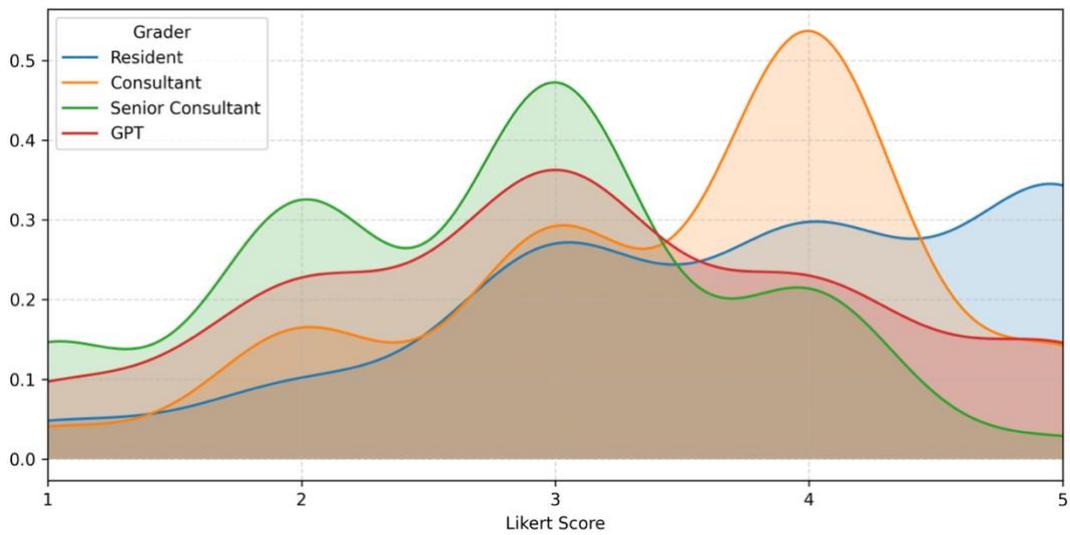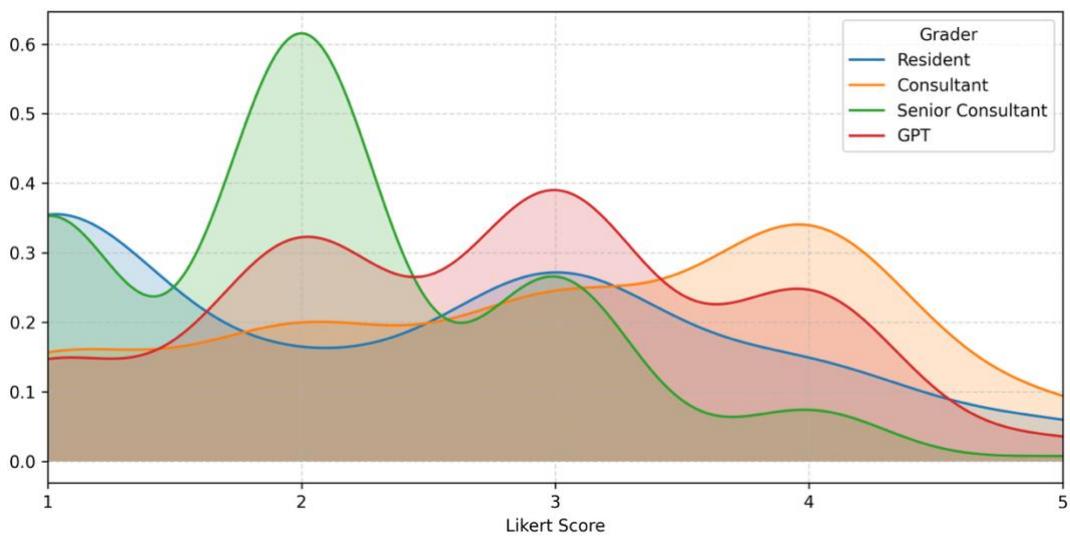

*Figure 4.* **Performance of Medical-based Large Language Models Across Ophthalmology Subspecialties based on aggregated evaluation of all graders**

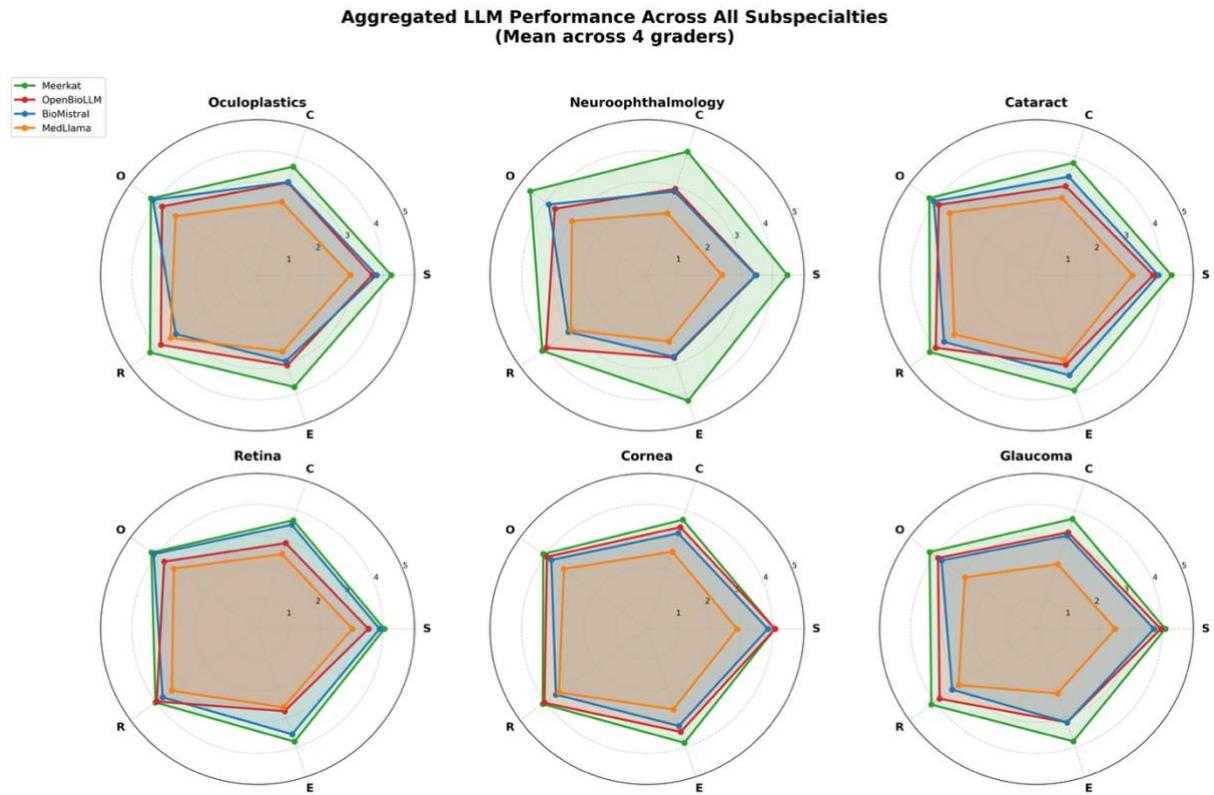

**Supplementary Materials**

*Supplementary Table 1.* Summary of the testing dataset consisting of 180 ophthalmology-related patient queries across 6 subspecialties

| Subspecialty | 180 Test questions |
|---|---|
| Neuroophthalmology | What are the investigations of myasthenia gravis? |
| Neuroophthalmology | What is the treatment for myasthenia gravis affecting the eyes? |
| Neuroophthalmology | What is the treatment for double vision? |
| Neuroophthalmology | What is the treatment for blepharospasm or hemifacial spasm? |
| Neuroophthalmology | What are the symptoms and signs of optic neuritis? |
| Neuroophthalmology | What are the side effects of botox injection to my eyelid? |
| Neuroophthalmology | What is migraine with visual aura? |
| Neuroophthalmology | What is the treatment of migraine with aura? |
| Neuroophthalmology | What is 6th nerve palsy? |
| Neuroophthalmology | What investigations do I need to do if I am found to have optic neuritis? |
| Neuroophthalmology | Is there any treatment for optic neuritis? |
| Neuroophthalmology | My doctor says that optic neuritis may be associated with multiple sclerosis. What is multiple sclerosis? |
| Neuroophthalmology | What is the treatment for migraine with visual aura? |
| Neuroophthalmology | How is cranial 6th nerve palsy treated? |
| Neuroophthalmology | What is 3rd nerve palsy? |
| Neuroophthalmology | Is twitching of the eyelid and face the same? |
| Neuroophthalmology | How is a humphrey visual field test performed? |
| Neuroophthalmology | How is a goldman visual field test performed? |
| Neuroophthalmology | What are the risks of intravenous methylprednisolone for optic neuritis? |
| Neuroophthalmology | Is seeing double images worrying? |
| Neuroophthalmology | Will my vision in my right eye improve with time after NAION? |
| Neuroophthalmology | Is there any treatment for NAION of the eye? |
| Neuroophthalmology | Are there any reasons that will increase my risk of having NAION in my eye? |
| Neuroophthalmology | My doctor says that I had NAION in my left eye. What does that mean? |
| Neuroophthalmology | What are causes of misalignment of the eyes? |
| Neuroophthalmology | What is myasthenia gravis of the eyes? |
| Neuroophthalmology | Are there early signs to suggest squint in my child I should look out for? |
| Neuroophthalmology | What is the treatment for squints in children? |
| Neuroophthalmology | What are causes of squints in children? |
| Neuroophthalmology | What causes in-turning of the eyelids? |
| Oculoplastics | What is epiblepharon? |
| Oculoplastics | How is chalazion treated? |
| Oculoplastics | What are the risks of surgery to correct droopy eyelids? |
| Oculoplastics | What are some problems that can happen from in-turning of the eyelids? |
| Oculoplastics | What is the cause of ptosis? |
| Oculoplastics | What are the treatment options for tear duct blockage? |
| Oculoplastics | What should I do if there is bleeding after DCR? |
| Oculoplastics | What is dermatochalasis? |
| Oculoplastics | What are the signs of Bells palsy? |
| Oculoplastics | What can cause the eyelids to not be able to close fully? |
| Oculoplastics | What is the treatment of in-turning of eyelids? |
| Oculoplastics | Should I be worried about tearing in my newborn baby? |
| Oculoplastics | What is the treatment if the eyelids are not able to close fully? |
| Oculoplastics | How is eyelid surgery done? |
| Oculoplastics | How can thyroid problems affect the eyes? |
| Oculoplastics | What are some symptoms that patients with thyroid eye disease will experience? |
| Oculoplastics | How is thyroid eye disease treated? |
| Oculoplastics | Why is there a silicon tube sticking out of the corner of my eye after tear duct surgery? |
| Oculoplastics | What is the risk of tear drainage blockage surgery? |
| Oculoplastics | What are causes of tear drainage blockage? |
| Oculoplastics | What are the symptoms of blockage in tear drainage? |
| Oculoplastics | How are droopy eyelids treated? |
| Oculoplastics | What is ptosis? |
| Oculoplastics | What can result in one side of the face to droop? |
| Oculoplastics | Why are my eyes tearing excessively? |
| Oculoplastics | What are some symptoms that may suggest an eye tumor? |
| Oculoplastics | What are some problems with a prosthetic eye that patients may face? |
| Oculoplastics | Does having a lump on the eyelid mean that I have eye cancer? |
| Oculoplastics | What are the complications of surgery for droopy eyelids? |
| Oculoplastics | What does it mean if my shingles rash affects the tip of my nose as well? |
| Cataract | What is toric intraocular lens for cataract surgery? |
| Cataract | What are the preparation steps needed before my cataract surgery? |
| Cataract | What is the difference between monofocal and multifocal intraocular lenses used in cataract surgery? |
| Cataract | What are the different types of cataracts and how do they affect my vision? |
| Cataract | What will the follow up reviews be like after my cataract surgery? |
| Cataract | Can't cataract surgery be done by laser? |
| Cataract | Why is my vision poor after cataract operation? |
| Cataract | What is a Yag capsulotomy and why do I need it? |
| Cataract | Can a cataract come back after surgery? |
| Cataract | What are some reasons why I have cataracts? |
| Cataract | How is cataract surgery being done? |
| Cataract | What can go wrong during cataract surgery? |
| Cataract | My doctor told me that I have to go for manual small incision cataract surgery as my cataract is too thick. Can you explain what that is? |
| Cataract | Can cataracts happen in children? |
| Cataract | How are childhood cataracts treated? |

| Category | Question |
|---|---|
| Cataract | What are possible causes of cataracts in children? |
| Cataract | What can I expect in the operating room during my cataract surgery? |
| Cataract | How do I know when it is time to go for cataract surgery? |
| Cataract | What are the different refraction targets after cataract surgery? |
| Cataract | What is the treatment of refractive surprise after cataract surgery? |
| Cataract | What is refractive surprise after cataract surgery? |
| Cataract | What are enhanced depth of focus intraocular lenses for cataract surgery? |
| Cataract | What is a piggyback lens for the eye? |
| Cataract | What will my eye doctor check for during my review on the first day after cataract surgery? |
| Cataract | What is the different about extracapsular cataract surgery? |
| Cataract | Why is it important to inform my eye doctor that I did LASIK previously prior to my cataract surgery? |
| Cataract | What are the do's and don'ts after my cataract surgery? |
| Cataract | What do I have to take note of before going for my biometry test before cataract surgery? |
| Cataract | What does it mean to have capsular block after cataract surgery? |
| Cataract | What will my eye doctor do if my eye pressure is high first day after cataract surgery? |
| Retina | How do I know if I have diabetic retinopathy? |
| Retina | What are the treatment options for diabetic retinopathy? |
| Retina | How does retinal detachment happen? |
| Retina | What is the treatment for a blood vein blockage in my eye? |
| Retina | My doctor told me there is blockage in my retinal vein, what does that mean? |
| Retina | Can I leave my epiretinal membrane alone? |
| Retina | Is there any risk for the FFA or ICG test? |
| Retina | What can go wrong with IVT injection? |
| Retina | How do I use an amsler grid? |
| Retina | Can surgery improve my vision if I have AMD? |
| Retina | What is an epiretinal membrane? |
| Retina | Are there further tests needed if I have retinal vein occlusion? |
| Retina | What is retinitis pigmentosa? |
| Retina | What does diabetic retinopathy mean? |
| Retina | What is PRP or pan retinal photocoagulation laser? |
| Retina | Are there risks of PRP or panretinal photocoagulation laser? |
| Retina | What is photodynamic therapy to the reitna? |
| Retina | How is an IVT injection performed? |
| Retina | What does uveitis mean? |
| Retina | What are some reasons that can cause uveitis? |
| Retina | How is uveitis being managed? |
| Retina | How is central serous chorioretinopathy treated? |
| Retina | My doctor says I have a blister in the center of my retina, called Central Serous Chorioretinopathy. What exactly is that? |
| Retina | What is the treatment for retinoblastoma? |
| Retina | When should you worry if your child may have retinoblastoma? |
| Retina | What is retinoblastoma? |
| Retina | What is retinopathy of prematurity? |
| Retina | Can retinopathy of prematurity be treated? |
| Retina | What is age related macular degeneration? |
| Retina | What is the treatment for retinal detachment? |
| Cornea | Can dry eyes be treated? |
| Cornea | What will you the doctor do during pterygium surgery? |
| Cornea | What is a pterygium? |
| Cornea | My doctor told me I have Fuchs endothelial dystrophy. How does surgery help my corneal edema? |
| Cornea | What is the reason that I have corneal edema? |
| Cornea | What is astigmatism? |
| Cornea | What will the doctor do to check if I have a corneal infection? |
| Cornea | What is the treatment needed for corneal infection? |
| Cornea | How is keratoconus treated? |
| Cornea | What is keratoconus? |
| Cornea | What is the treatment with atropine eye drops for myopia like? |
| Cornea | What are Orthokeratology (OrthoK) lenses? |
| Cornea | What is Relex smile? |
| Cornea | What are the steps of the LASIK laser procedure? |
| Cornea | How long will vison correction after LASIK last? |
| Cornea | What is LASIK Xtra? |
| Cornea | I am currently breastfeeding my 2 months old child, can I go for LASIK? |
| Cornea | What is the difference in recovery after LASEK vs LASIK? |
| Cornea | Can corneal ulcers be prevented? |
| Cornea | What may I expereince if I have corneal edema? |
| Cornea | What are dry eyes? |
| Cornea | What are some symptoms you may experience with dry eyes? |
| Cornea | My doctor says I have oily eyelids, what does that mean? |
| Cornea | What is the treatment of blepharitis? |
| Cornea | What are the things I should do after my corneal transplant surgery? |
| Cornea | Why is a corneal transplant surgery indicated? |
| Cornea | What are things to avoid after my corneal transplant surgery? |
| Cornea | What is corneal graft rejection? |
| Cornea | What are the different types of corneal transplant surgery? |
| Cornea | How does my doctor confirm that I have corneal swelling? |
| Glaucoma | What does glaucoma mean? |
| Glaucoma | How do I know if I have glaucoma? |
| Glaucoma | How do I confirm if I have glaucoma? |
| Glaucoma | What are the treatment options for glaucoma? |
| Glaucoma | What will the doctor do to test my visual field? |
| Glaucoma | What are the different eye drops used for glaucoma? |
| Glaucoma | What are the side effects of glaucoma eye drops? |
| Glaucoma | Why do I need glaucoma surgery? |
| Glaucoma | What are the risks of glaucoma surgery? |
| Glaucoma | What puts me at higher risk of glaucoma? |
| Glaucoma | How do lasers help in glaucoma? |
| Glaucoma | What is SLT in glaucoma? |
| Glaucoma | How is SLT performed in glaucoma eyes? |
| Glaucoma | What should I take note of after LPI procedure for glaucoma? |
| Glaucoma | When will you worry if a baby has glaucoma? |
| Glaucoma | Are there different types of glaucoma? |
| Glaucoma | What does it mean by open angle glaucoma? |
| Glaucoma | What does closed-angle glaucoma mean? |
| Glaucoma | What will my glaucoma doctor do when I go for follow up appointments? |
| Glaucoma | Can glaucoma be prevented? |
| Glaucoma | What is narrow drainage angle in the eye? |
| Glaucoma | Are narrow drainage angles of the eye treated by laser? |
| Glaucoma | How is LPI laser for glaucoma performed? |
| Glaucoma | What are some problems that can happen from LPI laser for glaucoma? |
| Glaucoma | How is glaucoma tube surgery performed? |
| Glaucoma | What is glaucoma tube surgery? |
| Glaucoma | When are risks that can happen with MIGS glaucoma surgery? |
| Glaucoma | What should I expect after trabeculectomy surgery? |
| Glaucoma | How is trabeculectomy glaucoma surgery done? |
| Glaucoma | Can I have glaucoma even if my eye pressure is not high? |

*Supplementary Table 2:* **Inter-rater reliability amongst all graders across the 4 tested large language models**

| LLM | Evaluation component | ICC_A1 | Mean_Pairwise_Quadratic_Kappa |
|---|---|---|---|
| **MedLLaMA3-v20** | Safety | 0.395 | 0.3932519222212850 |
| **MedLLaMA3-v20** | Context & Consensus | 0.526 | 0.5185275826585580 |
| **MedLLaMA3-v20** | Objectivity | 0.244 | 0.271505646858452 |
| **MedLLaMA3-v20** | Reproducibility | 0.393 | 0.3925348591968230 |
| **MedLLaMA3-v20** | Explainability | 0.473 | 0.4768378857626030 |
| **Meerkat-7B** | Safety | 0.339 | 0.3437861735285380 |
| **Meerkat-7B** | Context & Consensus | 0.385 | 0.37661886329193900 |
| **Meerkat-7B** | Objectivity | 0.176 | 0.23130096808840800 |
| **Meerkat-7B** | Reproducibility | 0.219 | 0.2181264277681800 |
| **Meerkat-7B** | Explainability | 0.290 | 0.29183814753715000 |
| **BioMistral-7B** | Safety | 0.254 | 0.2633657739373590 |
| **BioMistral-7B** | Context & Consensus | 0.348 | 0.34739837301585400 |
| **BioMistral-7B** | Objectivity | 0.107 | 0.13914978477870300 |
| **BioMistral-7B** | Reproducibility | 0.381 | 0.38310833993164000 |
| **BioMistral-7B** | Explainability | 0.317 | 0.3208364890843380 |
| **OpenBioLLM-8B** | Safety | 0.265 | 0.29241296845698900 |
| **OpenBioLLM-8B** | Context & Consensus | 0.327 | 0.32897525206569500 |
| **OpenBioLLM-8B** | Objectivity | 0.111 | 0.16986039173898100 |
| **OpenBioLLM-8B** | Reproducibility | 0.358 | 0.37339319665512600 |
| **OpenBioLLM-8B** | Explainability | 0.340 | 0.3556964795547140 |

*Supplementary Figure S1:* **Resident Grader Evaluation of LLM Performances across Ophthalmology Subspecialties**

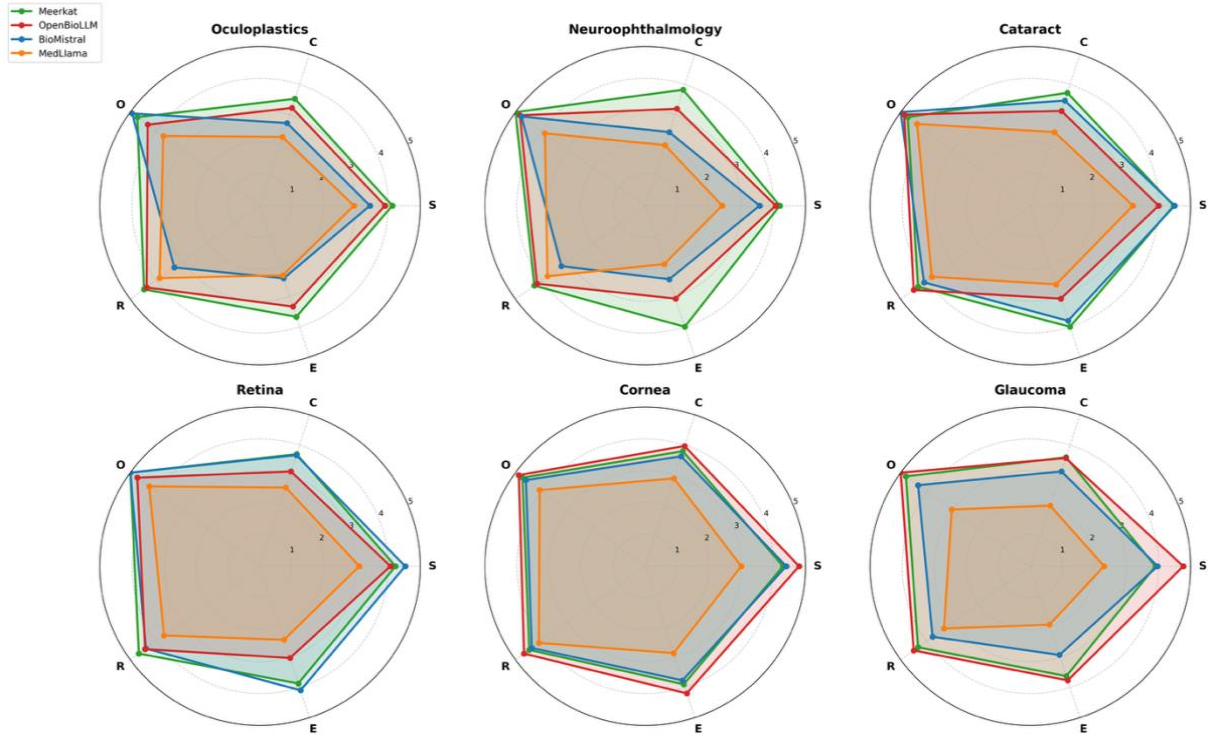

*Supplementary Figure S2:* **Consultant Grader Evaluation of LLM Performances across Ophthalmology Subspecialties**

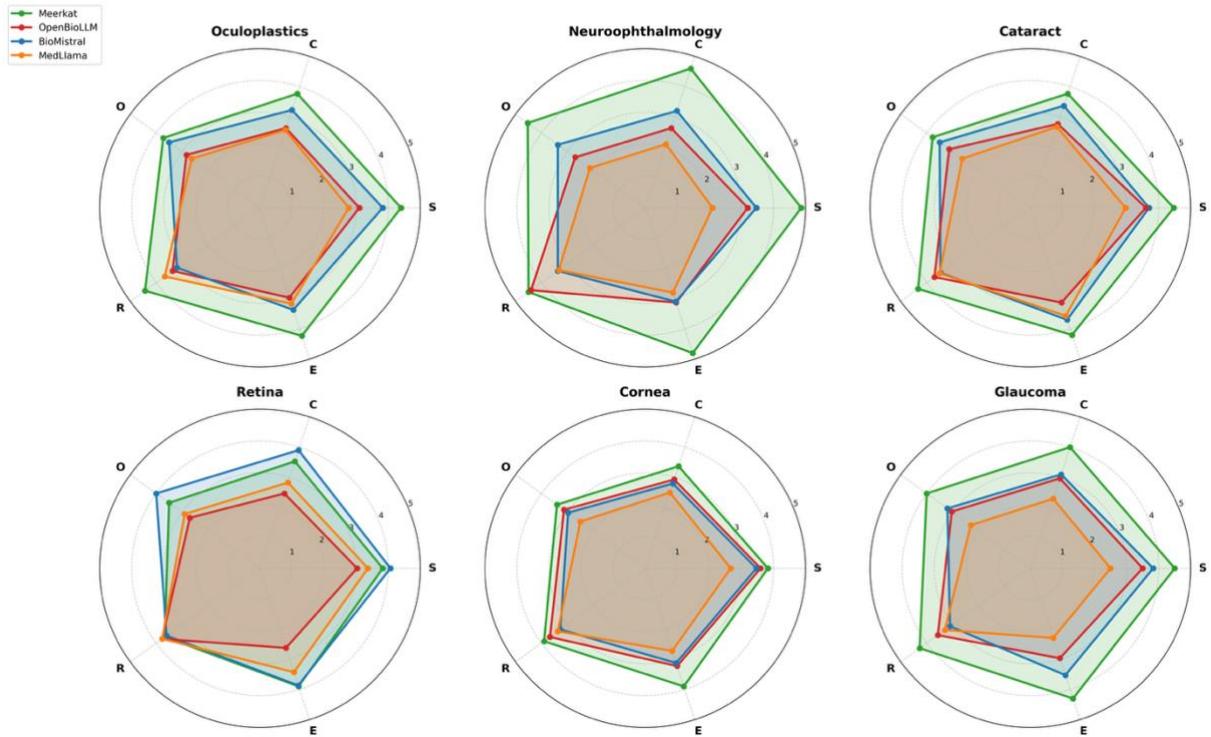

*Supplementary Figure S3:* **Senior Consultant Grader Evaluation of LLM Performances across Ophthalmology Subspecialties**

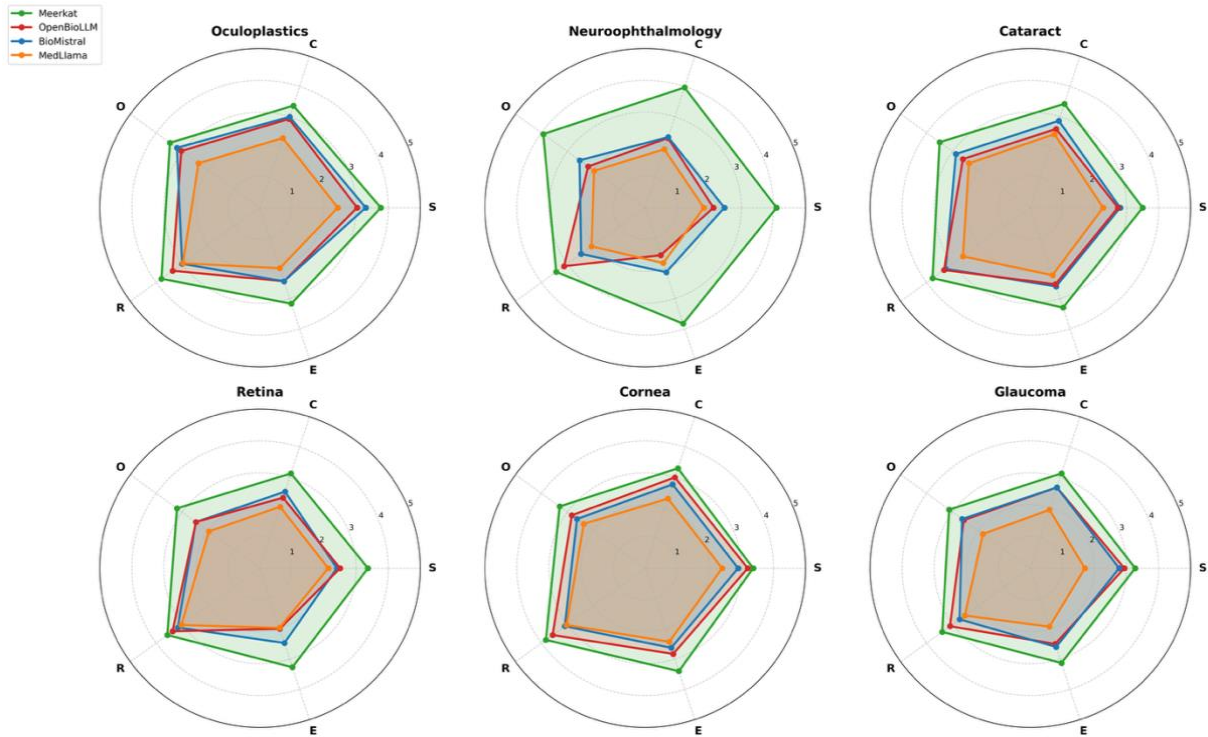

*Supplementary Figure S4:* GPT4-Turbo Grader Evaluation of LLM Performances across Ophthalmology Subspecialties

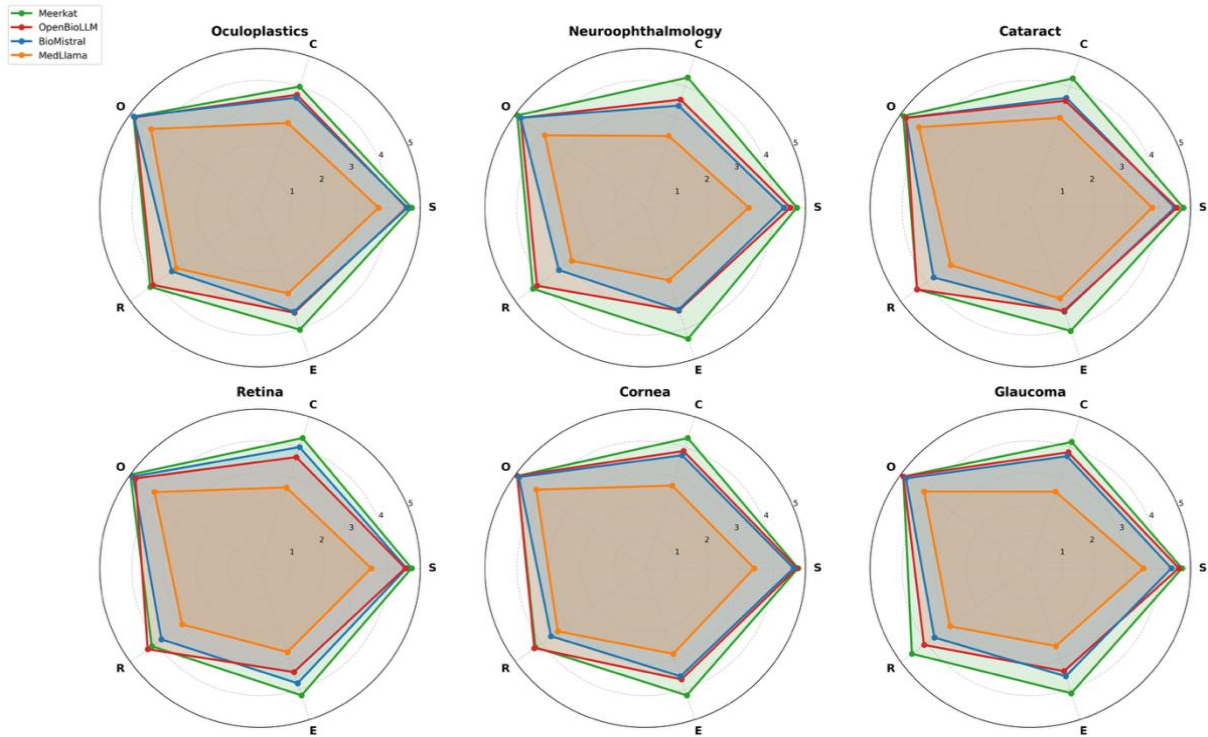

*Supplementary Figure S4:* Comprehensive Performance Matrix of Medical-based Large language models across Ophthalmology Subspecialties and Grader Expertise Levels

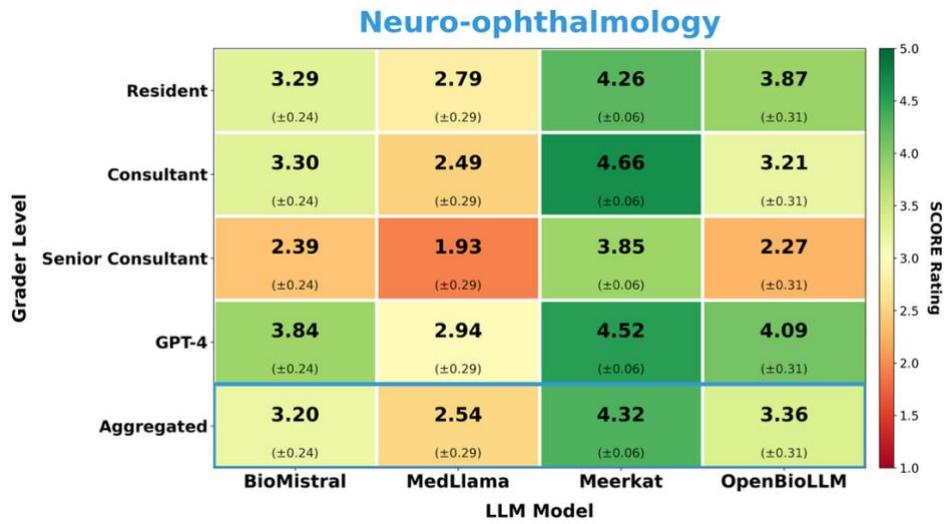

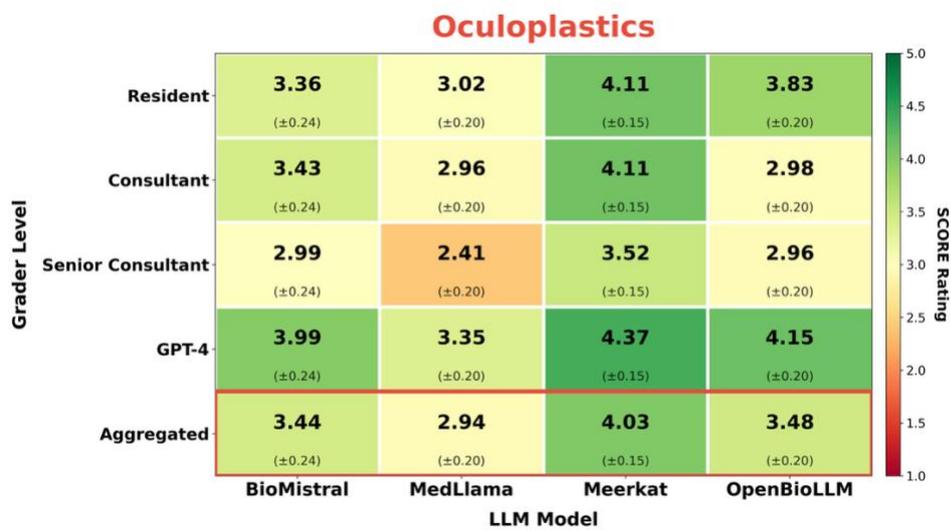

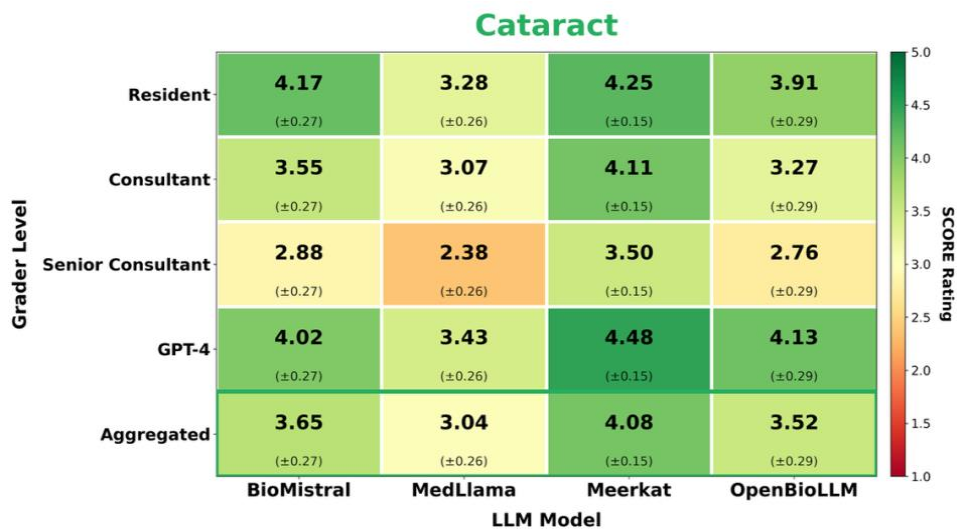

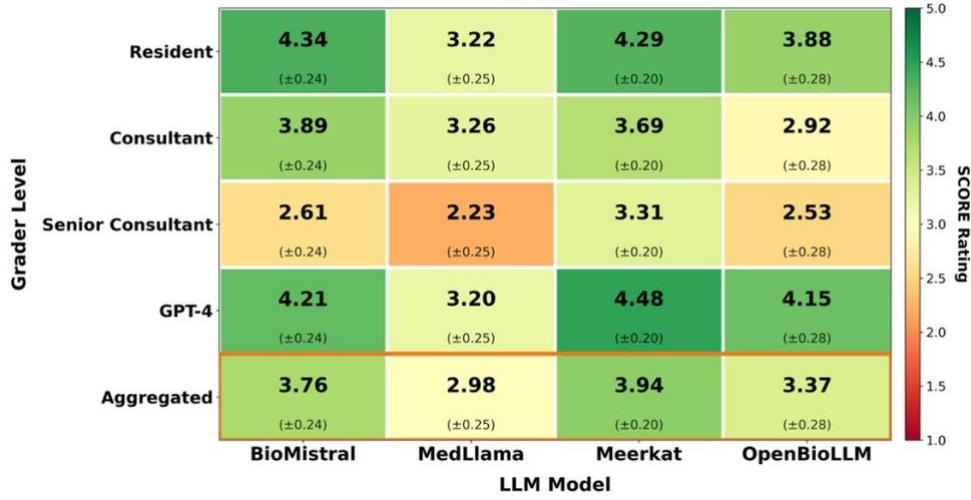
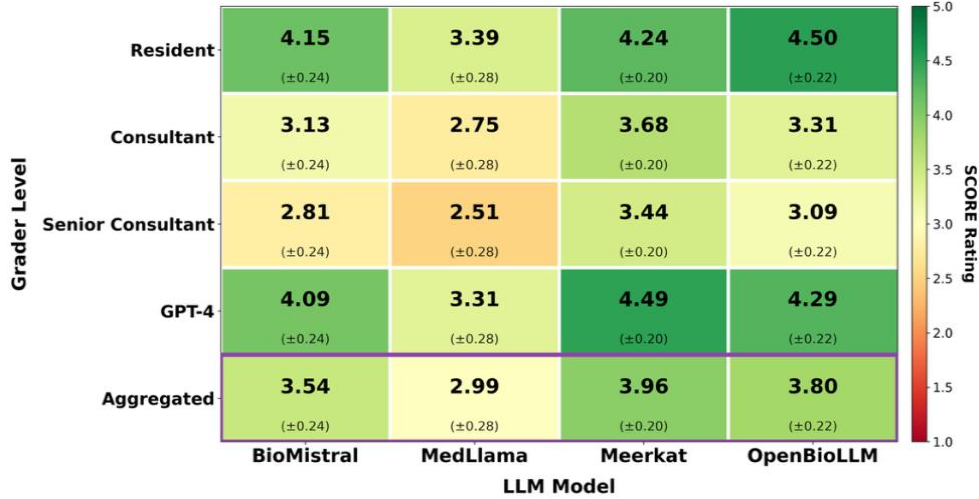
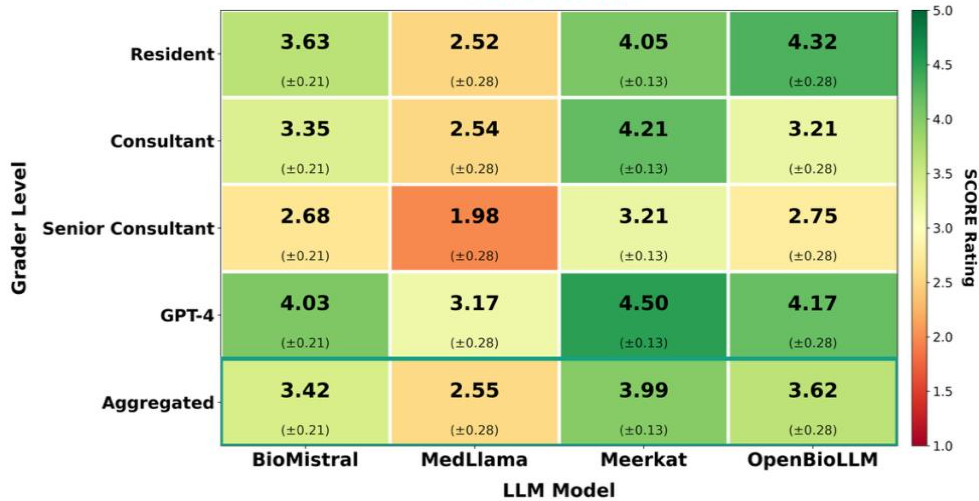